\documentclass[letterpaper, 10 pt, conference]{ieeeconf}  

\IEEEoverridecommandlockouts                              

\usepackage{soul}
\usepackage{amsmath,amssymb,amsfonts}
\usepackage{algorithmic}
\usepackage{graphicx}
\usepackage{tabularx}
\usepackage{textcomp}
\usepackage{xcolor}
\usepackage{subcaption}
\usepackage{url}
\usepackage{hyperref}
\usepackage{siunitx}

\usepackage[labelsep=period]{caption}
\renewcommand{\thetable}{\Roman{table}}

\title{\LARGE \bf
Towards Autonomous Pipeline Inspection with Hierarchical Reinforcement Learning}

\author{Nicolò Botteghi$^{1}$, Luuk Grefte$^{1}$, Mannes Poel$^{1}$, Beril Sirmacek$^{2}$, Christoph Brune$^{1}$, \\ Edwin Dertien$^{1}$, and Stefano Stramigioli$^{1}$
\thanks{$^{1}$Nicolò Botteghi is with Faculty of Electrical Engineering, Mathematics and Computer Science,
        University of Twente, Enschede, The Netherlands
        {\tt\small n.botteghi@utwente.nl}}%
\thanks{$^{2}$Beril Sirmacek with the Department of Computer Science, J\"{o}nk\"{o}ping University, J\"{o}nk\"{o}ping, Sweden}
}

\begin{document}

\maketitle
\thispagestyle{empty}
\pagestyle{empty}



\begin{abstract}
    Inspection and maintenance are two crucial aspects of industrial pipeline plants. While robotics has made tremendous progress in the mechanic design of in-pipe inspection robots, the autonomous control of such robots is still a big open challenge due to the high number of actuators and the complex manoeuvres required.
    To address this problem, we investigate the usage of Deep Reinforcement Learning for achieving autonomous navigation of in-pipe robots in pipeline networks with complex topologies.
    Moreover, we introduce a hierarchical policy decomposition based on Hierarchical Reinforcement Learning to learn robust high-level navigation skills.
    We show that the hierarchical structure introduced in the policy is fundamental for solving the navigation task through pipes and necessary for achieving navigation performances superior to human-level control. A video of our experiments can be found at: \url{https://youtu.be/uyjSHulpGoI}.
\end{abstract}



\section{INTRODUCTION}


Pipelines networks are the fulcrum of the oil and gas industries and of gas and water mains. These pipes must be periodically inspected to guarantee the safety and proper functioning of the plants. However, inspection is usually a long, expensive and tedious procedure that requires the shut-down of the whole plant and, in the specific case of industrial pipelines, the removal of the insulation around the pipes. With metal pipes, the inspection is currently performed from the outside using ultrasonic or magnetic probes that measure the wall thickness. Unfortunately, these inspection methods provide limited information about the state of pipes due to very low resolution and noisy data and require the removal of the insulation. 
Currently, Pipeline Inspection Gauges, or PIGs, are used to inspect the pipelines from the inside. However, while PIGs do not require a full shut down of the plant, they cannot be used to inspect networks with complex topologies, e.g. sharp corners, T-junctions and vertical sections. 

In the last two decades, inspection robotics has focused on designing new robotic prototypes for in-pipe inspection. However, especially in the case of small diameter pipes, the mechatronics of these robots is complex, costly and with low operability \cite{Abdellatif2018MechatronicsDO}. While the design of these in-pipe robots has quickly progressed, many steps have yet to be taken to navigate and inspect complex pipes autonomously. In-pipe inspection robots operate in highly constrained environments, with limited sensing equipment, without or with limited knowledge of the pipeline-network structures beforehand and unpredictable situations due to the contact dynamics and slippage as different fluids and media can be present in the pipes during the inspection. Furthermore, these robots are often composed of multiple joints, and multiple actuators have to be simultaneously controlled, making the design of an autonomous, robust and adaptable navigation system challenging. 

Reinforcement Learning \cite{sutton_reinforcement_2018}, or RL,  has proven to be a valuable solution for many robotics challenges and tasks such as mobile robot navigation, dexterous manipulation through robotic arms, and bipedal robot locomotion \cite{kober2013reinforcement}. However, when the task requires the execution of a sequence of complex skills on a long temporal horizon, Reinforcement Learning algorithms tend to struggle \cite{nachum2018data}.
Hierarchical Reinforcement Learning, or HRL, takes advantage of the hierarchical policy decomposition to exploit underlying problem structures and simplify the learning of complex tasks. The hierarchical decomposition can be either defined by using prior knowledge \cite{sutton1999between}, \cite{heess2016learning}, \cite{florensa2017stochastic}, \cite{Frans2017MetaLS}, or can be automatically learned during training \cite{nachum2018data}, \cite{vezhnevets2017feudal}, \cite{bacon2016optioncritic}. While the latter category of algorithm does not require expert knowledge for defining the hierarchy, the autonomous discovery of the options often leads to sub-optimal policies if additional regularizers are not used during the learning phase \cite{florensa2017stochastic}, \cite{bacon2016optioncritic}.

\begin{figure*}[ht!]
    \centering
    \includegraphics[width=0.55\linewidth,page=3]{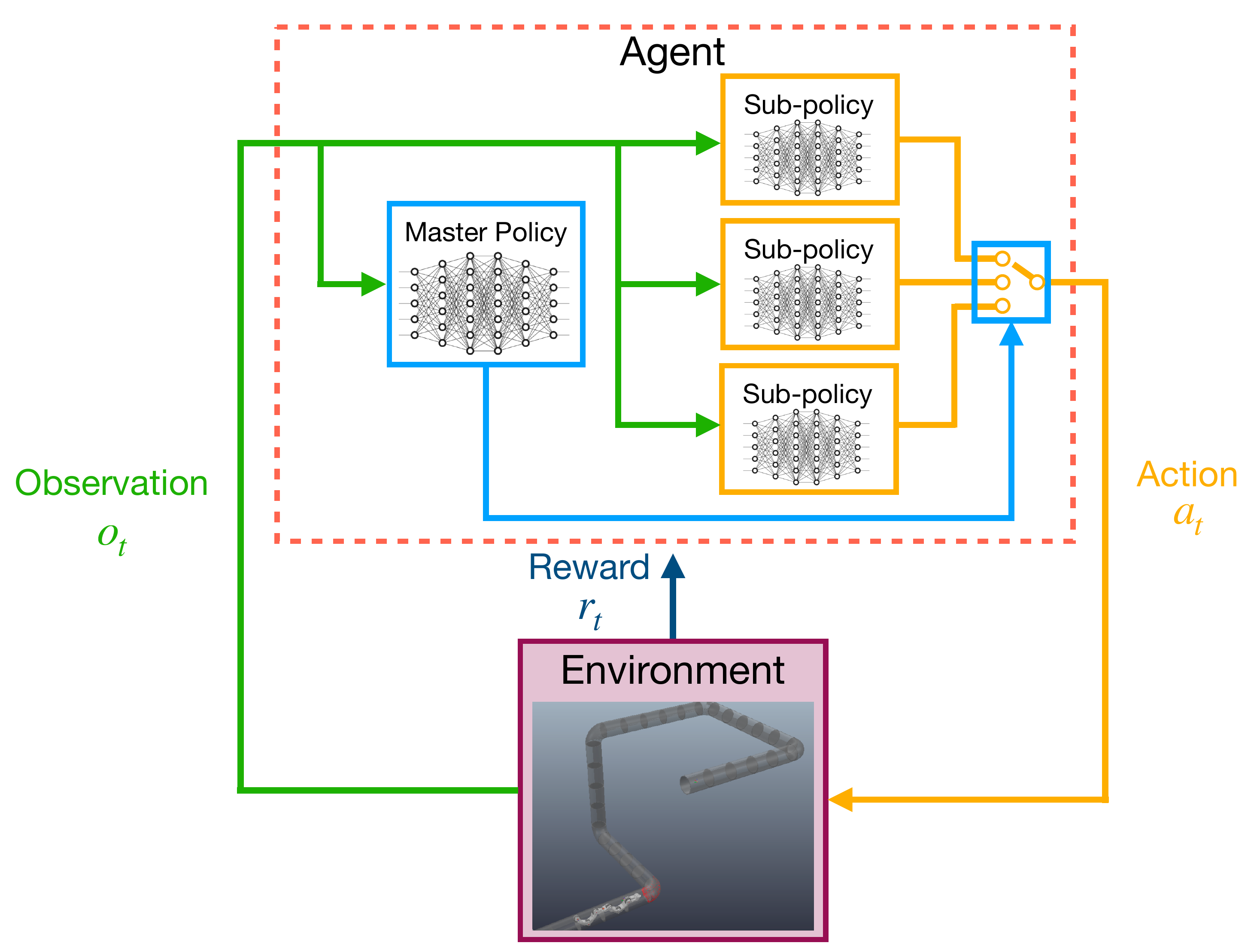}
    \caption{Proposed Hierarchical Reinforcement Learning architecture for the autonomous navigation of the PIRATE robot.}
    \label{fig:HRL_framework_intro}
\end{figure*}

We propose a Hierarchical Reinforcement Learning framework for autonomous navigation of multi-actuated, complex robots for in-pipe inspection. This is shown in Figure \ref{fig:HRL_framework_intro}. In particular, we focus on the pipe inspection robot called PIRATE  \cite{dertien2011development}, but the approach can be easily adapted to many different snake-like pipe inspection robots. The framework combines expert knowledge for determining the hierarchy of the policies with simple auxiliary reward functions for ensuring the optimal behaviour of the sub-policies.  To the best of the authors' knowledge, this is the first time a robust and flexible control solution employing Reinforcement Learning for the autonomous navigation of snake-like pipe inspection robots with a clamping mechanism is presented.
In this paper, we aim at addressing the following research questions:
\begin{enumerate}
    \item What is a good hierarchy for learning robust and generalizable policies for the pipeline inspection robot PIRATE?
    \item What are the benefits of such a hierarchical decomposition of the policy?
    \item How do the learned policies compare to human-expert control?
\end{enumerate}

The paper is organized as follows: Section \ref{sec:background} presents the theoretical background and Section \ref{sec:related work} briefly surveys the state of art of in-pipe inspection robotics. Then Section \ref{sec:methodology} explains the methodology followed in this research, followed by Section \ref{sec:experimental_design} showing the experimental design. Section \ref{sec:experimental_results} presents and discusses the results of the experiments. Eventually, Section \ref{sec:conclusions} concludes the paper.


\section{BACKGROUND} \label{sec:background}
\subsection{Reinforcement Learning}
Reinforcement Learning \cite{sutton_reinforcement_2018} is the Machine Learning branch dealing with the problem of optimal sequential decision-making. In the Reinforcement Learning framework, an agent tries to learn the optimal behavior for solving a given problem by interacting with an unknown environment. The interaction can be formally studied by means of Markov Decision Processes, or MDPs. An MDP is a tuple $\langle \mathcal{S}, \mathcal{A}, \text{T}, \text{R} \rangle$  where $\mathcal{S}$ corresponds to the set of states, $\mathcal{A}$ to the set of actions, $\text{T}: \mathcal{S} \times \mathcal{A} \longrightarrow [0, 1]$ to the transition function determining the evolution of the states and $\text{R}:\mathcal{S} \times \mathcal{A}\longrightarrow \mathbb{R}$ is the reward function. The goal of a Reinforcement Learning agent is the maximisation of the total cumulative reward $\sum_{t=0}
^{T} \lambda^t r_t$, where $r_t$ is the scalar reward obtained for taking the action $a_t$ in the state $s_t$ at time step $t$, and $\lambda \in [0, 1]$ is the discount factor weighting the importance of future rewards.

\subsubsection{Proximal Policy optimisation}

Proximal Policy optimisation, or PPO, \cite{Schulman2017ProximalPO} is an actor-critic policy gradient method that improves Trust Region Policy optimisation algorithm, or TRPO, \cite{schulman2015trust}, by relaxing its hard constraints. PPO replaces such constraints of TRPO by using loss functions for training the policy neural network. We use the variant of PPO with the clipped objective function, shown in Equation (\ref{PPOloss}), for its popularity, but the proposed framework is not strictly dependent on it.
\begin{equation}
\mathcal{L}_{clip}(\theta) = \mathbb{E}[\min(\zeta_t(\theta)\hat{A}_t, \textit{clip}(\zeta_t(\theta), 1-\epsilon,1+\epsilon)\hat{A}_t]
\label{PPOloss}
\end{equation}
where $\zeta_t(\theta)=\frac{\pi_{\theta}(a_t|s_t)}{\pi_{\theta old}(a_t|s_t)}$ corresponds to the probability ratio of the current policy and the old policy, $\hat{A}_t$ the estimation of the advantage function, $\epsilon$ is the clipping coefficient and $\theta$ the parameters' vector of the policy.

\subsection{Hierarchical Reinforcement Learning}
Hierarchical Reinforcement Learning algorithms aim at exploiting the structure of the problems by learning hierarchically-structured policies, efficiently solving complex tasks, and improving the generalization of the learned behaviors. In this context, the MDP model is often extended to Semi-Markov Decision Process, or SMDP, model \cite{howard1971dynamic} to take into account the temporal dimension introduced by the abstract actions. A SMDP is a tuple $\langle \mathcal{S}, \mathcal{A}_a, \mathcal{B}, \text{T}, \text{R} \rangle$ where $\mathcal{S}$ is the set of states, $\mathcal{A}_a$ is the set of abstract actions, $\mathcal{B}$ is the set of all possible duration of the abstract actions, $\text{T}: \mathcal{S} \times \mathcal{A}_a \times \mathcal{B} \longrightarrow [0, 1]$ is the transition function and $\text{R}:\mathcal{S} \times \mathcal{A}_a \times \mathcal{B}  \longrightarrow \mathbb{R}$ is the reward function.

\section{RELATED WORK} 

Exploiting structures and hierarchies is one of the most important challenges for scaling RL algorithms to more complex real-world problems. A famous Hierarchical Reinforcement Learning approach is the so-called Option framework \cite{sutton1999between}, \cite{stolle2002learningoptions}. Options are temporally extended abstract actions corresponding to the set of skills the agent need to learn in order to solve tasks. In this framework, above the options, we always find a high-level policy, learning to select the best option for the given context. Options can be either hand-crafted based on prior knowledge, as in \cite{heess2016learning}, \cite{florensa2017stochastic}, \cite{Frans2017MetaLS}, or can be automatically learned during training, as in the option-critic framework proposed by \cite{bacon2016optioncritic}. In the latter case, however, regularization, e.g. entropy maximisation, has to be often employed to prevent the learning of sub-optimal options.

A different approach is followed by HIRO \cite{nachum2018data}, and FuN \cite{vezhnevets2017feudal}, where the high-level policy does not select anymore which abstract action to actuate, but, instead, determines abstract goals for the low-level policies. Similar to the option framework, the abstract goals are chosen with a lower frequency than the actions chosen by the low-level policies.

In our work, we want to exploit the high amount of prior knowledge we have about the robot's mechanics and motion and about the navigation task in the structured pipeline networks. Therefore we find the option framework the most suitable for this scenario.



\section{AUTONOMOUS PIPELINE INSPECTION ROBOTS} \label{sec:related work}

In the last two decades, many innovative and different designs of in-pipe inspection robots have been proposed. According to \cite{surveypiperobots2017}, most of these robots use wheels for locomotion, have modular and snake-like bodies, and clamp inside the pipes. These three elements, when combined, allow the maximum flexibility of usage in different pipeline structures with vertical sections, junctions and corners.

To the category of modular and snake-like robots belongs MAKRO \cite{Rome1999TowardsAS}, a sewer inspection robot with an articulated body and multiple wheels. This robot, however, cannot clamp itself in the pipes, which significantly reduces the range of its usage.
Another example is the robot proposed in \cite{Choi2002RoboticSW} for inspection of urban gas pipelines. The robot has complicated mechanics with many joints and wheels, and it can clamp inside the pipe, allowing travelling even vertical sections.
In \cite{Selvarajan2019DesignAD} a bio-inspired snake robot is designed and presented. However, while the control principle is introduced, no actual test in complex pipes is shown.
Eventually, the PipeTron \cite{Debenest2014PipeTronS} and the PIRATE \cite{dertien2011development} are snake-like robots with wheel-based locomotion and the ability to clamp inside the pipes. These features allow these two robots to be very flexible in terms of the range of use and functionalities.

Differently from the previously cited works, KANTARO \cite{Nassiraei2007ConceptAD} employs wheels for locomotion. Its body is simple, allowing easier movements through junctions and reductions of the pipe diameters. The motion of this robot is, however, limited to planar pipeline networks without vertical sections. A similar design is used in \cite{Abdellatif2018MechatronicsDO} with the addition of a traction system for improving locomotion.

While the designs of these robots have greatly progressed in the past year, autonomous inspection is still a big open challenge \cite{Rome1999TowardsAS}. Especially for the complex snake-like robots with clamping mechanisms, the control and navigation are very complicated, and no fully autonomous and robust solutions for navigating complex pipeline networks with sharp corners and vertical sections are yet present.

\subsection{PIRATE Robot}

\begin{figure}[h!]
    \centering
    \includegraphics[width=0.75\linewidth]{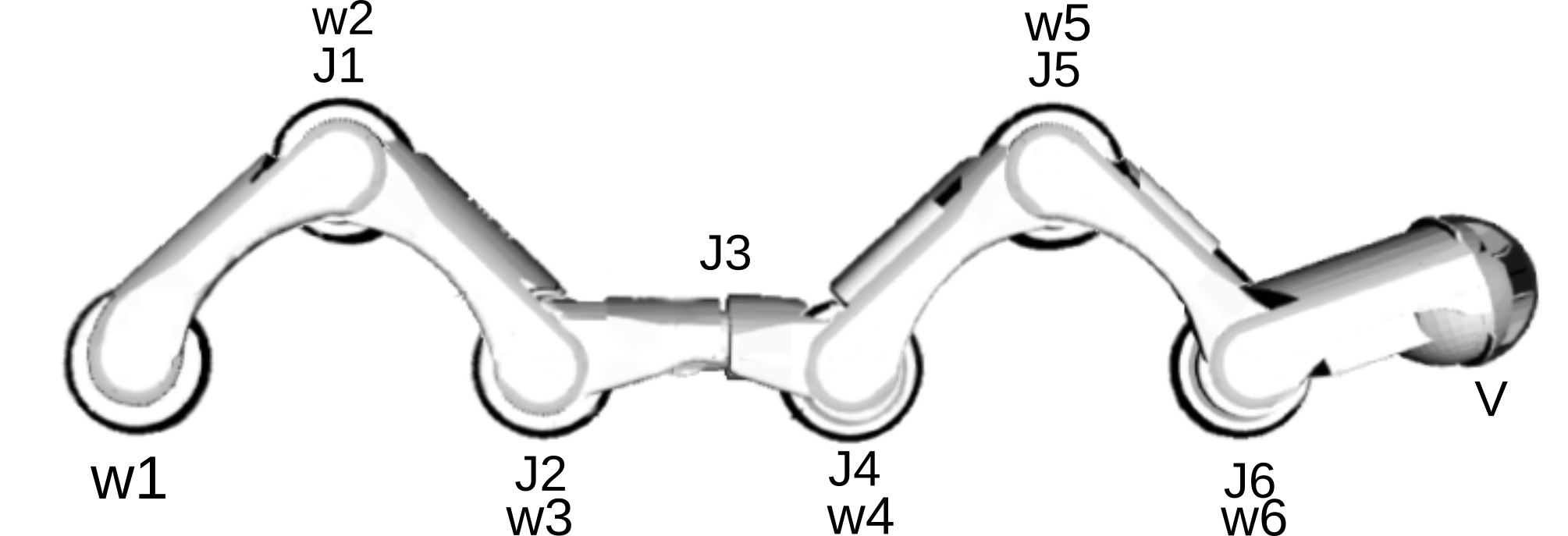}
    \caption{Schematic drawing of the PIRATE robot. $w_i$ indicates the $i^{th}$-wheels, $J_i$ the $i^{th}$-joints and $V$ the vision sensor. Figure reproduced and adapted from \cite{dertien2011development}.}
    \label{fig:schematic_pirate}
\end{figure}

The PIRATE robot \cite{dertien2011development} is designed to travel through pipes having different diameters, vertical sections and sharp corners. 
The PIRATE robot has six actuated joints $J_{1,\dots,6}$ and six actuated wheels $w_{1,\dots,6}$, as shown in Figure \ref{fig:schematic_pirate}. 
The robot is composed of two (inverted) V-shape sections ($w_{1, 2, 3}$ and $w_{4, 5, 6}$) that allow its clamping inside the pipes. Each joint and wheel can be either controlled through position or velocity set-points fed to low-level (PID) controllers running on embedded boards. The robot is equipped with absolute encoders for measuring the joints and wheels' rotation and with inertial measurement units (IMU) for measuring acceleration and orientation.  
Moreover, the robot perception is enhanced with cameras for visual inspection and with Light Detection and Ranging (LiDAR) sensors for navigation.



\section{METHODOLOGY} \label{sec:methodology}

\subsection{Reinforcement Learning for the PIRATE Robot} \label{subsec: state_action_space}



\subsubsection{The Action Space}
The action space $\mathcal{A} \in \mathbb{R}^{12}$ is chosen to be continuous to be able to execute smoother and more advanced manoeuvres. For this reason, the Reinforcement Learning algorithm chosen is PPO, similarly to \cite{Frans2017MetaLS}.
Each wheel $w_{1,\dots,6}$ and each clamping joint $J_{1,2,4,5}$ are controlled using velocity commands, while the rotational joint $J_3$ and the vision-sensor joint $J_6$ are controlled using position commands, as shown in Table \ref{tab_1}. Our experiments have found beneficial the position control of the rotation joint $J_3$ and the vision-sensor joint $J_6$ for achieving more accurate motion. These two joints are critical for the orientation procedure of the robot with respect to corners\footnote{Due to its mechanics, the PIRATE robot requires a specific relative orientation with respect to the pipe corners in order to navigate through it.}, and for their detection, respectively. 
 
\begin{table} [h!]
\centering
\scalebox{1.0}{
\begin{tabular}{ ||c|c|c|| } 
 \hline
 Control Mode & \textit{velocity} & \textit{position} \\
 \hline
 Action space & $J_{1,2,4,5}, w_{1, \dots, 6} $& $J_3, J_6$ \\
\hline
\end{tabular}}
\caption{Control modes for the PPO policy, \textit{velocity} corresponds to the velocity control of the actuators, while \textit{position} to the position control.}
\label{tab_1}
\end{table} 
 
\subsubsection{The State Space}
We compare and analyse two alternative state spaces: the \textit{kinematic} and the \textit{visual}. The \textit{kinematic} state-set  includes all the kinematic information such as the positions $q_{J_{1,\dots,6}}$, orientations $o_{J_{1,\dots,6}}$ and velocities $v_{J_{1,\dots,6}}$ of the joints in the space, the positions $p_{w_{1,\dots,6}}$ and velocities $v_{w_{1,\dots,6}}$ of the wheels in the space, and the previous actions taken $a_{t-1}$. The \textit{visual} state-set extends the \textit{kinematic} one with external perception, i.e. 3D depth images $i_d$.

It is worth mentioning that the absolute positions, orientations of the robot in the space can be estimated, for example, by means of any Simultaneous Localization and Mapping algorithm \cite{thrun2002probabilistic}. However, for the sake of simplicity, we assume to have a good estimate available. Moreover, wheels and joints relative positions and velocities can be measured using the onboard sensors (encoders and IMUs).

\subsubsection{The Reward Function}

For a generic inspection task, the goal is to inspect as much of the pipeline as possible. Thus the reward function that the agent tries to maximise is proportional to the forward distance travelled by the robot. The reward function is shown in Equation (\ref{reward_distance}).
\begin{equation}
    \text{R}(s_t, a_t) = d_t - d_{t-1}
    \label{reward_distance}
\end{equation}
where $d_t$ corresponds to the absolute position of the robot at time-step $t$ and $d_{t-1}$ to the position at time-step $t-1$ with respect to the origin of the chosen reference frame.  This reward function encourages the agent to drive the robot forward in the pipeline. 

\subsection{Hierarchical RL and Policy Decomposition} \label{sec:HRL}

\begin{figure}[h!]
    \centering
\includegraphics[width=1.0\linewidth]{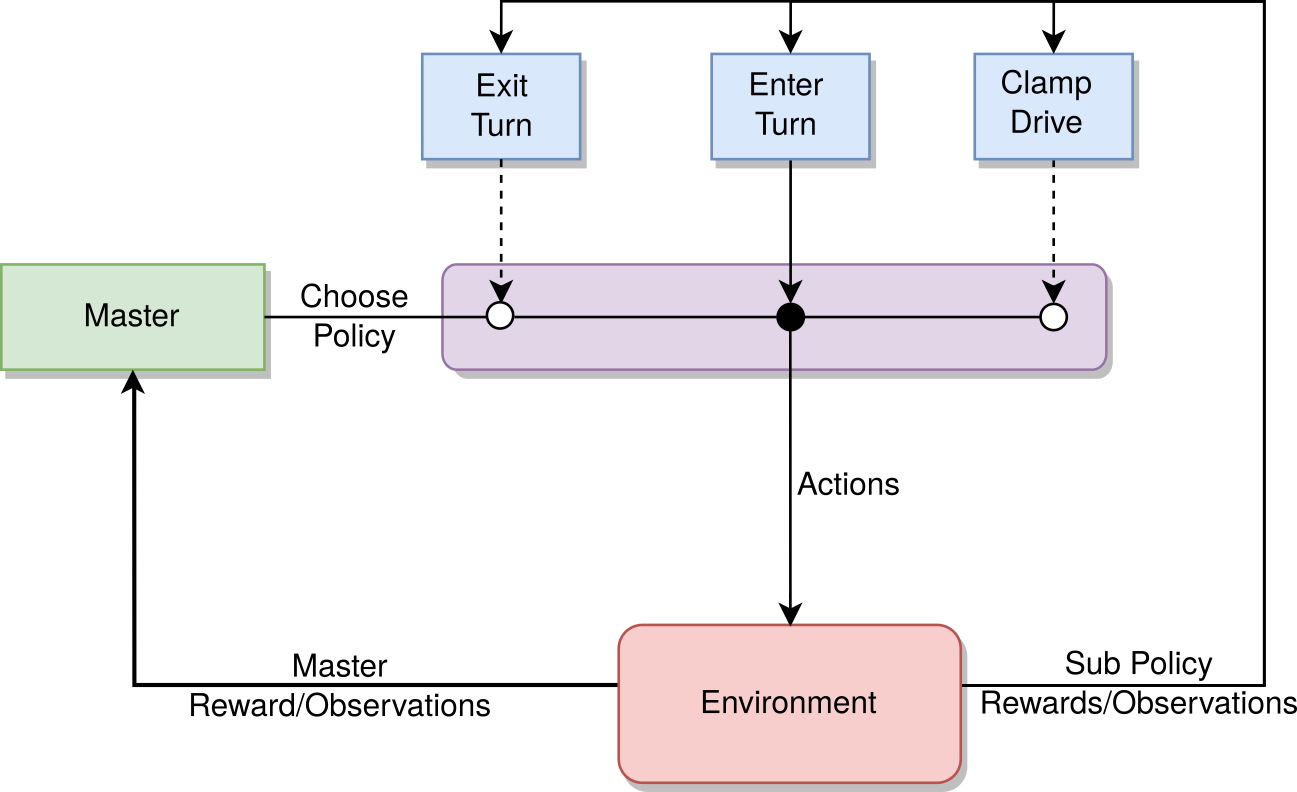}
    \caption{Hierarchical decomposition of the policies.}
    \label{fig:hrl_policy}
\end{figure}

While the task of learning to navigate might be easy for differential-drive mobile robots, for complex multi-actuated robots in constraint environments, this is not the case. In particular, moving through a straight section or a corner of the pipeline networks requires complicated coordination of actuators over long sequences of atomic actions. However, the high-level control is relatively straightforward, and all pipe inspection robots share the same needs of either travelling through straight pipes or moving through corners and junctions.  
We can exploit such a structure by hierarchically decomposing the Reinforcement Learning policy into a \textit{Master} policy and three \textit{sub-policies}.

\subsubsection{Master Policy and Sub-Policies}

We introduce a two-level hierarchical policy architecture, inspired by the Option framework \cite{sutton1999between}, where the \textit{Master} policy chooses which of the three sub-policies to deploy to maximise the reward function, in Equation (\ref{reward_distance}). Each sub-policy corresponds to a specific skill the agent requires to navigate through the pipeline networks, namely clamping and driving, and entering and exiting a corner, as shown in Figure \ref{fig:hrl_policy}.

\begin{figure}[!h]
\centering
\begin{subfigure}{0.45\textwidth}
  \centering
  \includegraphics[width=1.0\textwidth]{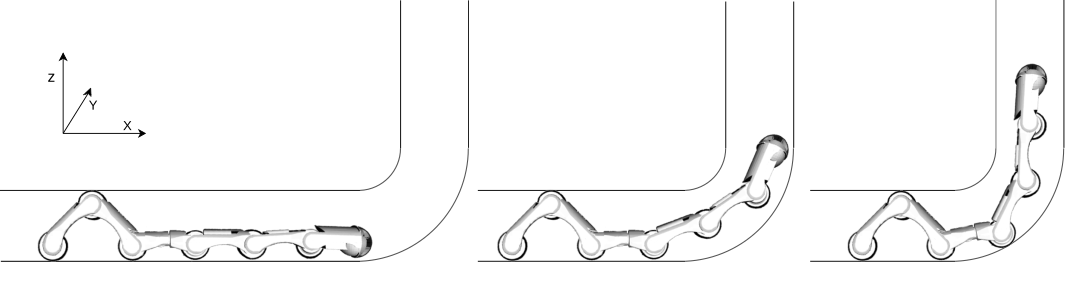}
  \captionsetup{justification=centering}
  \caption{\textit{Enter Turn}}
  \label{fig:enter}
\end{subfigure}
\begin{subfigure}{0.45\textwidth}
  \centering
  \includegraphics[width=0.70\textwidth]{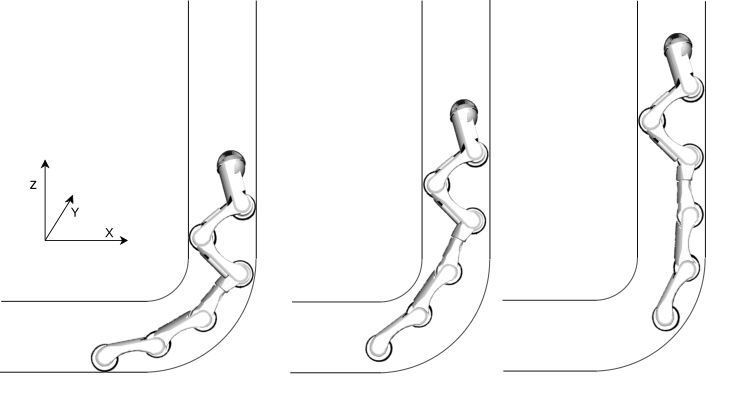}
  \captionsetup{justification=centering}
  \caption{\textit{Exit Turn}}
  \label{fig:outward}
  \end{subfigure}
\captionsetup{justification=centering}
\caption{Motion of the PIRATE robot through a corner.} 
\label{fig:turning_subpolicies}
\end{figure}

The \textit{Clamp Drive} sub-policy is responsible for clamping the robot in the pipes and drive through straight sections. Due to the robot's mechanics, driving and clamping are dependent actions, and the robot cannot travel any pipe without first clamping its body in it. The \textit{Enter Turn} sub-policy is in charge of driving the first V-shape of the robot through the corner, while \textit{Exit Turn} completes the turning procedure and clamps the robot in the new pipe segment, as shown in Figure \ref{fig:turning_subpolicies}. The turning procedure is the most challenging to learn. Thus we split this task into two different skills.

The \textit{Master} policy tries to learn to choose the correct abstract action, or skill, for the circumstance. This policy receives observation only when a specific sub-policy has finished its execution, i.e. after 30 time-steps or when an early stopping condition is met (e.g. when the goal is reached) in our case. Because the \textit{Master} policy picks action with a lower frequency than the sub-policies, it is even more important that such policy receives good observations from the environment. Otherwise, it is not able to choose the best sub-policy to enable. For example, if the depth camera is pointing at the ceiling or the bottom of the pipe, and no clear view of the pipe is present, the \textit{Master} policy has no way to know if the robot is close to a corner and what is its orientation. However, the hierarchical decomposition of the task allows us to define auxiliary reward functions for each of the sub-policies that can speed up the learning of the skills and consequently of the task. This is discussed in Section \ref{subsubsec:rew_aux}.

Another advantage of the policy decomposition is the possibility to constrain the action spaces of each sub-policy. This decomposition, again, allows learning better policies more efficiently. The choice of the different action spaces is shown in Table \ref{tab_2}. While the \textit{Clamp Drive} policy keeps the action space defined in Table \ref{tab_1}, while the \textit{Enter Turn} and \textit{Exit Turn} disable the front-wheels rotation and back-wheel rotation respectively. These wheels are not needed for moving through corners and junctions. Moreover, the control mode of joints is switched to position control, except for joints $J_1$ and $J_5$ respectively\footnote{The joints $J_1$ and $J_5$ need to hold the clamping of their V-shapes of the robot, and in our experiments, we have discovered velocity control mode optimal for such task.}, for more accurate manoeuvring. 

\begin{table} [h!]
\centering
\scalebox{1.0}{
\begin{tabular}{ ||c|c|c|c|| } 
 \hline
 Control Mode & \textit{velocity} & \textit{position} & \textit{N/A} \\
 \hline
 \textit{Clamp Drive} & $J_{1,2,4,5}, w_{1, \dots, 6} $& $J_3, J_6$ & - \\
 \textit{Enter Turn} & $J_1, w_{1,2,3}$ & $J_{2,\dots,6}$ & $w_{4,5,6}$\\
 \textit{Exit Turn} & $J_5, ,w_{4,5,6}$ & $J_{1,2,3,4,6}$& $w_{1,2,3}$ \\
\hline
\end{tabular}}
\caption{Action space and control modes for the different sub-policies. \textit{velocity} corresponds to velocity control of the actuators, \textit{position} to position control and \textit{N/A} if the actuator is not used by the sub-policy.}
\label{tab_2}
\end{table}

This choice of hierarchy is driven by the nature of the task, the robot's knowledge, and the topology of the pipeline networks. However, it can be easily adapted to different pipe inspection robots, given the similarities of tasks and mechanics of such robots. The three sub-policies define three different high-level skills the agent has to learn to navigate any pipe. In general, one could think of lower-level sets of skills, such as, for example, two independent clamping and unclamping policies for the V-shapes or one policy for actuating all the wheels independently of the joints. However, this would be detrimental to the overall performance. Due to the robot mechanics, most of the low-level atomic skills are dependent on each other, e.g. if both V-shapes are clamped, the PIRATE robot cannot actuate the joint $J_3$. Thus the \textit{Master} policy would need to learn an even more complex sequence of commands. Moreover, with a finer discretisation of the skills, the \textit{Master} policy task would become more complex as it is harder to distinguish which skill to use in each state. The proposed hierarchy trades off the task-complexity for the \textit{Master} policy and the sub-policy.

\subsubsection{Auxiliary Reward Functions for the Sub-Policies}\label{subsubsec:rew_aux}

To specialise the sub-policy and to quickly learn the skills, we define a \textit{clamping} reward that promotes clamping. This is particularly useful in the  \textit{Clamp Drive} policy. Additionally, we employ a \textit{depth-camera} reward for promoting the proper orientation of the depth camera in the direction of the pipe axis to obtain more and better information about the pipe network.
 
\section{Experimental Design} \label{sec:experimental_design}

\begin{figure*}[!ht]
\centering
\begin{subfigure}{0.49\textwidth}
  \centering
  \includegraphics[width=0.9\textwidth]{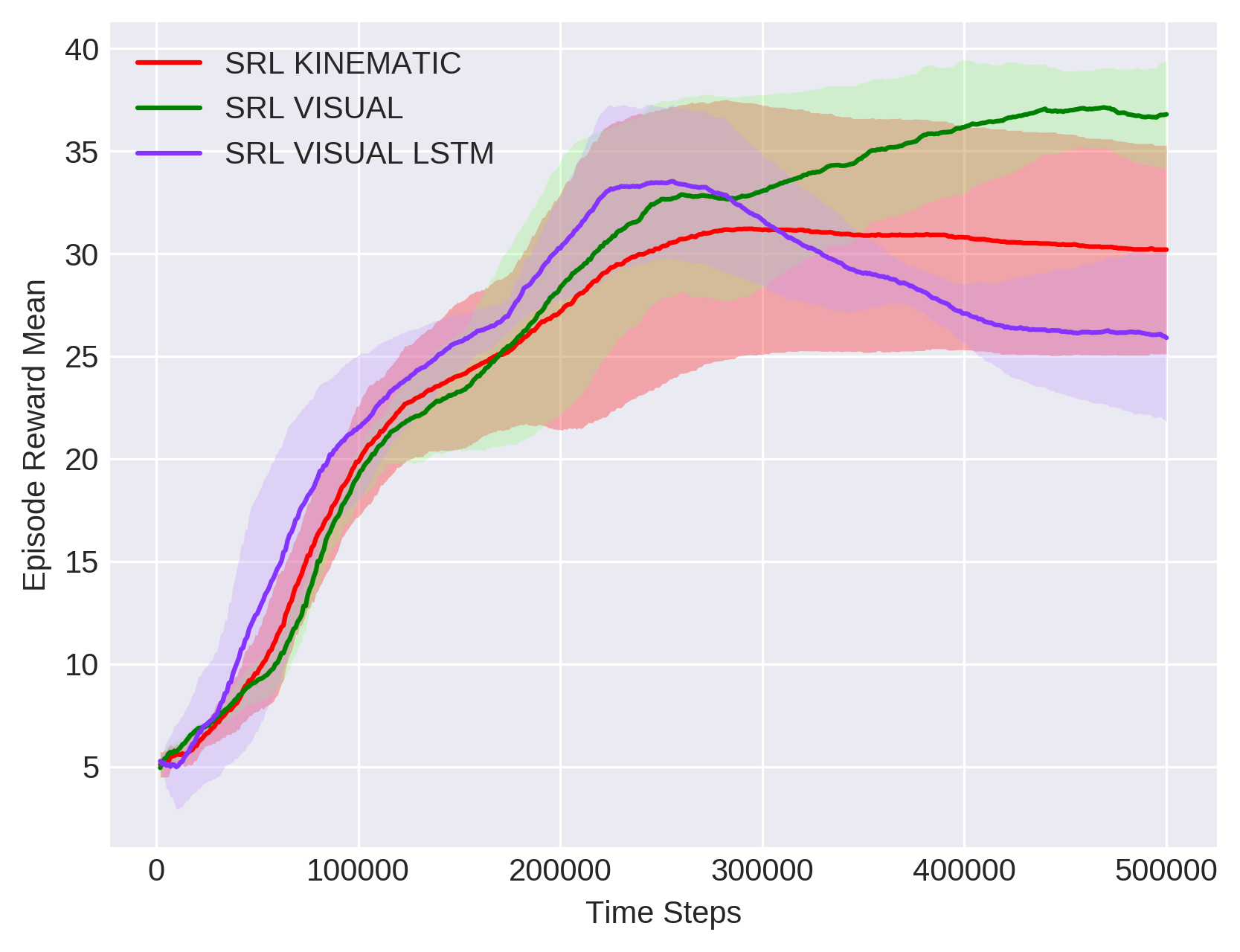}
  \captionsetup{justification=centering}
  \caption{}
  \label{fig:singleRLstatic}
\end{subfigure}
\begin{subfigure}{0.49\textwidth}
  \centering
  \includegraphics[width=0.9\textwidth]{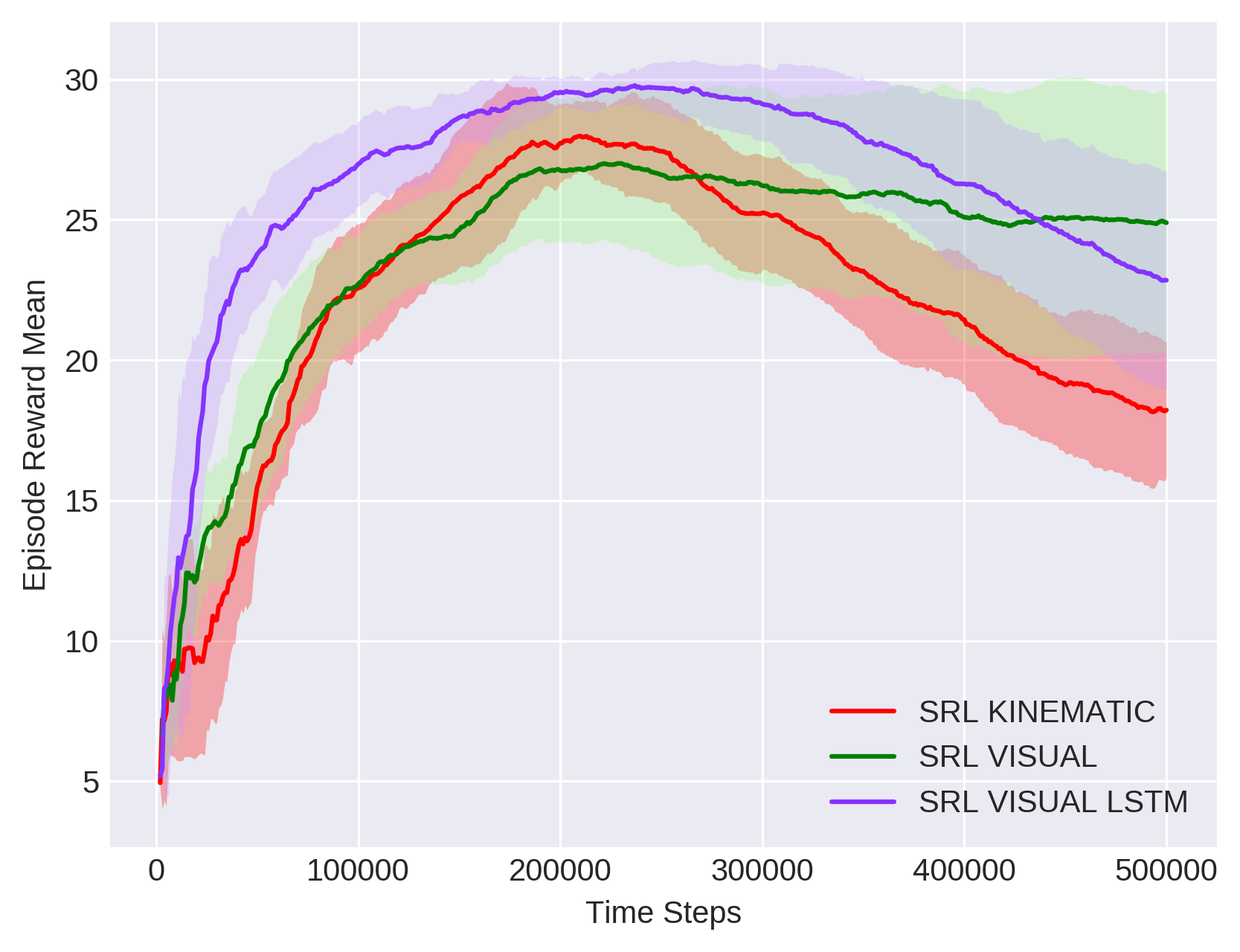}
  \captionsetup{justification=centering}
  \caption{}
  \label{fig:singleRLdynamic}
  \end{subfigure}
\captionsetup{justification=centering}
\caption{Cumulative reward during training of the PPO policy with different observation groups in the case of \textit{static} environment (Figure \ref{fig:singleRLstatic}) and \textit{dynamic} (Figure \ref{fig:singleRLdynamic}). The solid line represents the average cumulative reward, while the shaded area its standard deviation.} 
\label{fig:singleRLobs}
\end{figure*}

\subsection{Autonomous Navigation in Pipeline Networks}

The goal of the experiments is to learn an optimal navigation policy for manoeuvring the PIRATE robot in an unknown a priori pipeline network. In this work, we focus on the navigation problem in pipeline networks composed of straight sections and \SI{90}{\degree} bends with different smoothness, either 1D and 2D\footnote{Pipe corners are classified by indicating the smoothness of the corner with respect to the diameter of the pipe D}. We assume that the agent does not know the position of the end-point of the inspection mission, but it can detect it once the robot reaches it. This scenario represents a generic inspection mission. The robot has to autonomously navigate until a problem, e.g. a crack or corrosion in the pipe wall and a consequent reduction of the wall thickness, is found.

We train and test the agents in a \textit{static} pipeline, i.e. the pipeline is not changing over training nor testing. In a \textit{dynamic} pipeline, i.e. the pipeline network is randomly changing configuration in each training episode. The possible changes occurring in the pipeline are changes in the direction of the corners and relative distance among them.

\subsection{Simulation Environment}

We test the approach in the simulation environment V-REP \cite{coppeliaSim} using a model of the PIRATE robot. V-REP allows for realistic physical simulations and allows learning transferrable policies to real robots \cite{james2017transferring}. Bullet physical engine \cite{erickson2019assistive} is used.  Moreover, the Ray framework and the RLlib library \cite{moritz_ray_2018} are used for developing the HRL framework. OpenAI Gym  \cite{1606.01540} support is available in RLlib, and this is used to build the RL environment. OpenAI Gym provides a framework to build a custom RL environment. RLlib communicates with V-REP by using PyRep \cite{james2019pyrep}. Furthermore, RLlib uses the TensorFlow library for building neural network models. The PPO implementation of RLlib is used for training the agents. 


\subsection{State Space and Neural Network Architectures}

We first study the effect of the two different state-sets, introduced in Section \ref{subsec: state_action_space}, on the performances of a single PPO agent, with action space defined in Table \ref{tab_1}, when trained on a \textit{static} and a \textit{dynamic} pipe configuration.  

\subsubsection{Network Architecture}

In the case of the \textit{kinematic} state-set, the policy and the value function networks share a single fully connected layer of dimension 128 with tanh activation. The features are then fed to two identical branches composed of two fully connected layers of dimensions 128 and 64, respectively, with tanh activations. The output layer for the policy network has the dimension of the action space, while it has dimension 1 for the value function network.

When the \textit{visual} state-set is employed, the depth images are initially separated from the kinematic information of the state vector and pre-processed by two convolutional layers, with 5 and 10 filters, respectively, of size $5 \times 5$, and stride 2, and by a fully-connected layer. These features, extracted from the depth images, are then concatenated to the other elements of the state vector and fed to the same architecture used by the \textit{kinematic} state-set. A similar architecture is successfully employed in \cite{james2017transferring}. 

In the case of the \textit{visual} state-set, we also study the use of a recurrent architecture, adding a single LSTM layer to the \textit{visual} architecture, after the concatenation of the features from the depth images and the other components of the state vector. 


\subsection{RL vs HRL}
To assess the value of the hierarchical policy architecture, we compare the single PPO agent with the HRL agent trained proposed hierarchical approach on both \textit{static} and \textit{dynamic} pipeline networks using the \textit{visual} observation group.

\subsubsection{Training Regime of the HRL Method} \label{sec:trainingRegime}
We test two different training approaches for the HRL architecture: simultaneous optimisation of the four policies and independent optimisation of the policies. In simultaneous optimisation, the \textit{Master} policy and the sub-policies are optimised at the same time in the same environment. In contrast, in the independent optimisation procedure, the sub-policies are first trained in specialised environments\footnote{For example, a specialised environment for the \textit{Clamp Drive} sub-policy is composed of only straight pipes.} and, only once optimised, the master policy is trained. 

\section{RESULTS AND DISCUSSIONS} \label{sec:experimental_results}

\begin{figure*}[ht]
\centering
\begin{subfigure}{0.45\textwidth}
  \centering
  \includegraphics[width=1\textwidth]{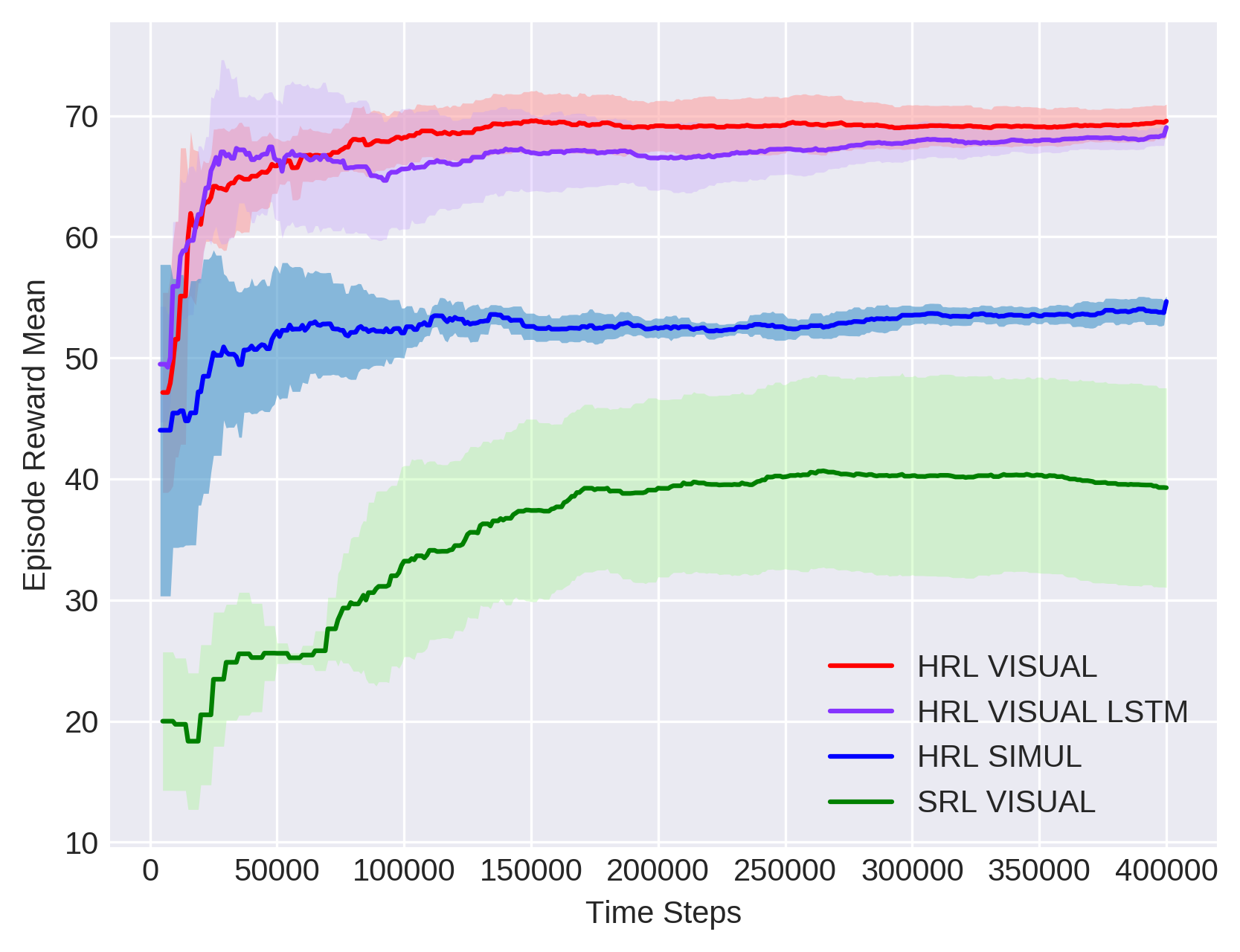}
  \captionsetup{justification=centering}
  \caption{}
  \label{fig:results_comparison}
\end{subfigure}
\begin{subfigure}{0.45\textwidth}
  \centering
  \includegraphics[width=0.95\textwidth]{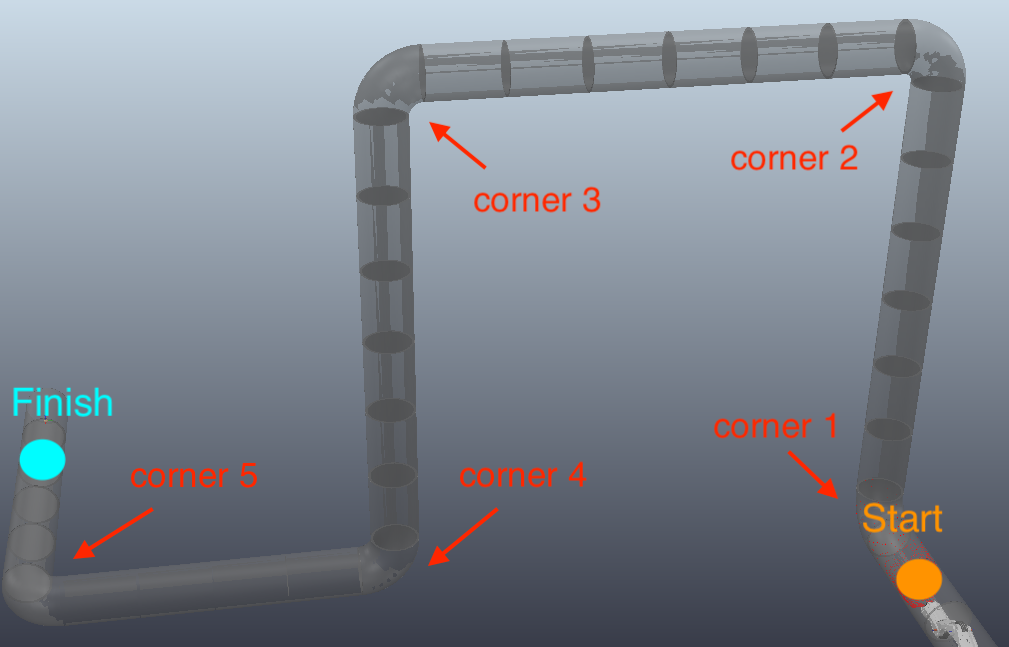}
  \captionsetup{justification=centering}
  \caption{}
  \label{fig:eval}
  \end{subfigure}
\captionsetup{justification=centering}
\caption{Comparison of HRL with sub-policy pre-training (red), HRL with joint optimisation of the sub-policies (blue) and single PPO agent (green).
The training results are shown in Figure \ref{fig:results_comparison} where
the average reward (solid line) and the standard deviation (shaded area) are presented. The evaluation pipeline network is shown in Figure \ref{fig:eval}.}
\label{fig:HRL-final}
\end{figure*}

\subsection{Different Observation Groups}

We first present the results obtained with the single PPO agent (SRL) with the two different observation groups in the \textit{static} and \textit{dynamic} environments. For each observation group, four independent experiments are performed, and the results are shown in Figure \ref{fig:singleRLobs}.

While the \textit{visual} state set is sufficient for completing the task\footnote{Considering the length of the pipes, a cumulative reward value above 35 means that the task is solved.} of navigating a 3-junctions pipe in the \textit{static} environment, see Figure \ref{fig:singleRLstatic}, it is not enough in the \textit{dynamic} case in which the three junctions randomly change position and orientation each training episode, as shown in Figure \ref{fig:singleRLdynamic}. The agent observing the depth images outperforms the one relying only on the kinematic information. Thus, we can conclude that it is not possible to learn a policy that drives the robot through the pipes by only looking at the kinematic information.
Moreover, in the \textit{dynamic} case, the \textit{kinematic} state-set generates catastrophic forgetting, as shown in Figure \ref{fig:singleRLdynamic}, where the performances of the policy start decreasing the more the training goes on. Surprisingly, in the \textit{dynamic} environment with only three junctions, even the \textit{visual} state-set does not allow sufficient policy improvements to reach the end of the pipe. The single PPO agent cannot learn a robust policy to deal with a random pipe-network configuration. Eventually, the \textit{visual} policy with LSTM does not improve the performances either in the \textit{static} case or in the \textit{dynamic} one. By including information on the positions and velocities of all the wheels and joints and depth information about the pipe network, the state vector already contains all the information needed to determine the state of the environment and, consequently, choose the best actions.

\subsection{Comparison of the RL and HRL Performances}

We now present the results obtained using the HRL framework proposed in Section \ref{sec:HRL} in a \textit{dynamic} 5-junctions pipe network. In particular, we compare HRL with pre-training of the sub-policies in different and specialised environments (HRL VISUAL) with the joint optimisation of all the policies (HRL SIMUL), as described in Section \ref{sec:trainingRegime}, and the single PPO (SRL VISUAL). The \textit{visual} state-set is used by all the agents. The average cumulative rewards obtained during training in the 5-junctions \textit{dynamic} environment are shown in Figure \ref{fig:results_comparison}. Again, four independent experiments are performed for each curve in the plot. Both HRL agents outperform PPO, but while HRL VISUAL consistently solves the task, 
HRL SIMUL struggles to achieve the same performance. The independent training of the sub-policies has a positive influence on the overall performances of the HRL-agent.

After the training phase, the performances of the agents are evaluated in unseen a priori environments. In Figure \ref{fig:eval}, we show the evaluation in a challenging and randomly generated pipeline with five junctions. We report in Table \ref{tab:compareresults} the distance travelled in meters by the different agents.

The single PPO agent (SRL VISUAL) can steer the robot only through the first corner and cannot advance more. The second corner requires the use of the rotation joint ($J_3$) to orient the front V-shape of the robot in the correct direction, but the PPO agent struggles to learn that. The proper re-orientation of the front is found one of the critical and most challenging manoeuvres for advancing through the pipe corners. The HRL agents perform better than PPO, but while the HRL with pre-training (HRL VISUAL) can steer the robot through all the five junctions, the HRL with simultaneous optimisation (HRL SIMUL) cannot make the robot escaping the fourth corner. The fourth corner requires great precision and robustness for each sub-policy, as it requires the re-orientation of the front modules of the robot while travelling upside down. This manoeuvre is more challenging to achieve when jointly training all the policies together. Again the policy with LSTM does not seem to improve the performance compared to the non-recurrent policy.
A video of our experiments can be found at: \url{https://youtu.be/H9IxZ1NYga4}. 



\subsection{Comparison with Human Control}

The PIRATE robot can be driven through the pipes by a human operator \cite{dertien2011development}, but only in the case of transparent pipes that allow the operator to see the robot constantly. 
However, this operation is not practical in real scenarios when the operators can only rely on onboard sensors to control the robot. 

To show the benefits of the use of Reinforcement Learning for controlling such pipe inspection robots, we try to manually control the robot in the same task of navigating through the pipe in Figure \ref{fig:eval}. For fairness of comparison, the operator cannot see through the pipes and can rely on the same action space of the RL agents (Table \ref{tab_1}) and the information contained in the \textit{visual} state-set, i.e. depth-camera images and kinematic information. 

When relying only on depth-camera images and kinematic information, a human struggles to complete the 5-junctions task, in Figure \ref{fig:eval}, even after many trials. In Table \ref{tab:compareresults} we record the best trails.
It is worth mentioning that a human operator may eventually solve the task for a fixed pipeline configuration. However, in the general inspection settings, the pipe configuration can change depending on the inspection task and can be even more complex. Thus, the human control of the robot does not seem a viable solution in this context.

\begin{table}[h!]
\centering
\scalebox{1.0}{
\begin{tabular}{||c|c||}
\hline
Method & Dist. traveled (m) \\
\hline
SRL VISUAL & 2.02 \\
HRL SIMUL & 4.83 \\
HRL VISUAL & $6.60^*$ \\
HUMAN & 4.88 \\ 
\hline
\end{tabular}}
\caption{Distance travelled in meters by different approaches in the evaluation environment in Figure \ref{fig:eval}. Only the HRL with pre-trained sub-policies (HRL VISUAL) completes the task.}
\label{tab:compareresults}
\end{table}



 
\subsection{Transferring the Policies to the Real Robot}
Because the experiments are conducted in simulated environments, the transfer of the learned policies to the real robot is an important aspect that has to be further investigated in future work. 

The proposed HRL framework has proven to be robust against perturbations of the observations, e.g. additive noise on sensory readings, and can generalize to different and untrained pipe configurations.
Moreover, while earlier research showed that policies, learned in virtual environments, relying on RGB images, are troublesome to transfer to the real-world \cite{matas2018simtoreal}, in our experiments, the agents only use a depth camera to perceive the environment and onboard sensory readings (encoders and IMUs) to obtain information of the robot's kinematic. Eventually, the agents' actions are position and velocity set-points for the robot's actuators. These set-points are then fed to low-level controllers that are in charge of tracking such reference values. By avoiding direct torque control of the actuators and by assuming well-tuned low-level controllers with similar performances to the one present on the actual robot, the difference in the dynamics between simulation and real-world is mitigated \cite{kober2013reinforcement}. 


\section{CONCLUSIONS} \label{sec:conclusions}
We presented a two-layers HRL framework for autonomous navigation of the snake-like pipe inspection robot PIRATE in pipeline networks with different topologies. 
We showed that the proposed hierarchical decomposition of the policies is necessary for solving the navigation problem of such robots in pipes with multiple junctions and with random configurations. 
The hierarchical decomposition allows learning a robust set of skills, and it is crucial for the generalization of the policy to unseen a priori pipeline networks. Moreover, with the proposed decomposition, the \textit{Master} policy can learn different inspection missions if retrained with a different reward function and without the need of retraining the sub-policies. 
Eventually, when the sub-policies are pre-trained separately in specialised environments, the hierarchical framework outperforms human-controlled operations.

\section*{APPENDIX}

\subsection{Hyperparameters Tuning}
Three independent grid-search experiments are performed to find a good set of hyperparameters. This is shown in Table \ref{tab-hyper}.

\begin{table} [h!]
\centering
\scalebox{1.0}{
\begin{tabular}{ ||c|| } 
 \hline
 Experiment 1 \\
 \hline
 Learning rate  \\
 {[}1e-4,1e-5*,1e-6{]}  \\
\hline
Experiment 2 \\
\hline
Clip, Lambda, Entropy-coefficient \\
{[}0.1,0.2*{]}, {[}0.99*,0.95{]} , {[}0.005*,0.01{]}  \\
\hline
Experiment 3 \\
\hline
Train-batch-size, Mini-batch-size \\
{[}500,1000*,2000{]}, {[}300*,100{]} \\
\hline
\end{tabular}}
\caption{Hyperparameters ranges used in the grid search experiments. * indicates the value used in the experiments presented in the paper.}
\label{tab-hyper}
\end{table}


\section*{ACKNOWLEDGMENT}
Nicolò Botteghi has received funding from Smart Tooling. Smart Tooling is an Interreg Flanders-Netherlands project sponsored by the European Union focused on automation in the process industry: making maintenance safer, cheaper, cleaner, and more efficient by developing new robot prototypes and tools.

\end{document}